\documentclass{article}
\usepackage{ijcai19}

\usepackage{times}
\usepackage{soul}
\usepackage{url}
\usepackage[draft]{hyperref}
\usepackage[utf8]{inputenc}
\usepackage[small]{caption}
\usepackage{graphicx}
\usepackage{amsmath}
\usepackage{booktabs}
\usepackage{algorithm}
\usepackage[skip=0pt]{caption}
\usepackage{caption}
\usepackage{amsfonts}
\usepackage{stmaryrd}
\usepackage{mathrsfs}
\usepackage{bm}
\usepackage{amsthm}

\usepackage{algpseudocode}
\usepackage[T1]{fontenc} %for bold font in xelex
\usepackage{enumerate}
\newtheorem{theorem}{Theorem}

\usepackage[titletoc]{appendix}

\title{Structured  Discriminative Tensor Dictionary Learning for \\Unsupervised Domain Adaptation}

\author{
Songsong Wu$^1$\and
Yan Yan$^{2}$\and
Hao Tang$^3$\and
Jianjun Qian$^4$\and
Jian Zhang$^5$\and
Xiao-Yuan Jing$^6$
\affiliations
$^1$Nanjing University of Posts and Telecommunications,
$^2$Texas State University\\
$^3$University of Trento,
$^4$Nanjing University of Science and Technology\\
$^5$University of Technology Sydney,
$^6$Wuhan University
\emails
sswu@njupt.edu.cn,
tom$\_$yan@txstate.edu,
hao.tang@unitin.it,
csjqian@njust.edu.cn,
Jian.Zhang@uts.edu.au,
jingxy$\_$2000@126.com
}
\begin{document}

\maketitle

\begin{abstract}
%	$\\$ [Background] $\\$
	Unsupervised Domain Adaptation (UDA) addresses the problem of performance degradation due to domain shift between training and testing sets, which is common in computer vision applications. Most existing UDA approaches are based on vector-form data although the typical format of data or features in visual applications is multi-dimensional tensor. Besides, current methods, including the deep network approaches, assume that abundant labeled source samples are provided for training. However, the number of labeled source samples are always limited due to expensive annotation cost in practice, making sub-optimal performance been observed. In this paper, we propose to seek discriminative representation for multi-dimensional data by learning a structured dictionary in tensor space. The dictionary separates domain-specific information and class-specific information to guarantee the representation robust to domains. In addition, a pseudo-label estimation scheme is developed to combine with discriminant analysis in the algorithm iteration for avoiding the external classifier design.	We perform extensive results on different datasets with limited source samples. Experimental results demonstrates that the proposed method outperforms the state-of-the-art approaches.
\end{abstract}

% \vspace{-0.1cm}
\section{Introduction}
% \vspace{-0.1cm}
A typical assumption in learning based visual recognition is that training and test data obey an identical distribution as they belong to the same domain. In practical applications, this assumption can be easily violated due to the distribution divergence of training data from source domain and test data from target domain. Such domain shift \cite{BDavid_ML10} is an universal issue in applications such as image recognition with varying lighting conditions and shooting angles of camera, challenging traditional recognition models. Domain adaptation \cite{SPan_TKDE10} addresses this issue by training the model using the data from both domains so as to transfer the discriminative knowledge from the source and the target.
% \begin{figure}[t] \small
%     \setlength{\abovecaptionskip}{-5pt}
% 	\centering
% 	\includegraphics[width=1\linewidth]{Fig_SDTDL.pdf}
% 	\caption{Illustration of SDTDL. $\mathbf{U}_{s}$, $\mathbf{U}_{t}$ and $\mathbf{W}_{c}$ characterise source domain, target domain and class semantics respectively. For $\mathcal{X}^c$ of class $c$, its domain factor and class factor are characterised separately by $\mathcal{A}^{c}_0$ and $\mathcal{A}^{c}_{c}$. Similarly, $\mathcal{B}^{c}_{0}$ and $\mathcal{B}^{c}_{c}$ compose the sparse representation of $\mathcal{Y}^{c}$. (Better view zoomed in and with color)}
% 		\label{fig:SDTDL}
% 		\vspace{-0.4cm}
% \end{figure}

Based on the amount of available labeled samples in target domain,
domain adaptation can be performed in two scenarios~\cite{Patel_SPM15}, semi-supervised domain adaptation (SDA) and unsupervised domain adaptation (UDA). In SDA, a small number of target samples are with class label, so it's essential to learn the discriminative model with the assistance of labeled source samples. The labels are unavailable in UDA, thus it relies on modeling the distribution relation between domains to achieve cross-domain recognition. In this paper, we aims to tackle domain shift problem in the scenario of UDA, which is more challenging and widespread in reality.

Instance adaptation \cite{YMansour_NIPS2009,YYu_ICML12} specifies the important weights of source samples in the objective function to match the data distribution of source and target domain. This principle works well only when the support of target distribution is contained in that of source distribution.
Feature adaptation seeks domain-invariant representations of samples so that their distributions are coincident and the discriminative information is preserved. The domain-invariant feature can be obtained through linear projection \cite{BFernando_ICCV13,BSun_AAAI16}, kernel mapping \cite{BGong_CVPR12,ZZhang_CVPR18},
%sparse coding \cite{MLong_CVPR13,Shekhar_CVPR13,BYang_AAAI18},
sparse coding \cite{Shekhar_CVPR13,BYang_AAAI18,tang2018deep},
and metric learning \cite{BKulis_CVPR11,SHerath_CVPR17}.
Classifier adaptation retrains a predefined classifier by learning the classifier parameters to guarantee its good generalization in the target domain \cite{LDuan_ICML09,RXu_CVPR18}.
Besides the aforementioned shallow learning based domain adaptation methods, domain adaptation via deep learning
\cite{YGanin_ICML15,MLong_ICML15,BKonstantinos_NIPS16,ETzeng_CVPR17,KSaito_CVPR18,JHoffman_ICML18} achieves
notable improvement and becomes increasingly popular. The deep DA methods extract nonlinear domain-invariant feature and train domain-robust classifier in an end-to-end manner.

Most shallow UDA methods treat data as vectors, meaning that multi-dimensional data such as images and videos or their features need to be converted from other form to vectors beforehand. This operation can incur several obstacles to domain adaption, including (1) the vectorization breaks the internal structure of data, which is demonstrated to be essential for recognition \cite{Aja_Book09}; (2) the vectorization increases the risk of model over-fitting because resulted vector is always long. Deep learning based domain adaptation methods encounter the dilemma of structure information loss because feature maps from convolutional layers need to be converted into vectors before they feed into the fully connected layers. In addition, the number of parameters in fully connected layers becomes large when feature map is transformed from tensor to vector, increasing the over-fitting risk of deep model, especially when training data are insufficient.

\begin{figure}[t]\small
	\centering
	\includegraphics[width=1\linewidth]{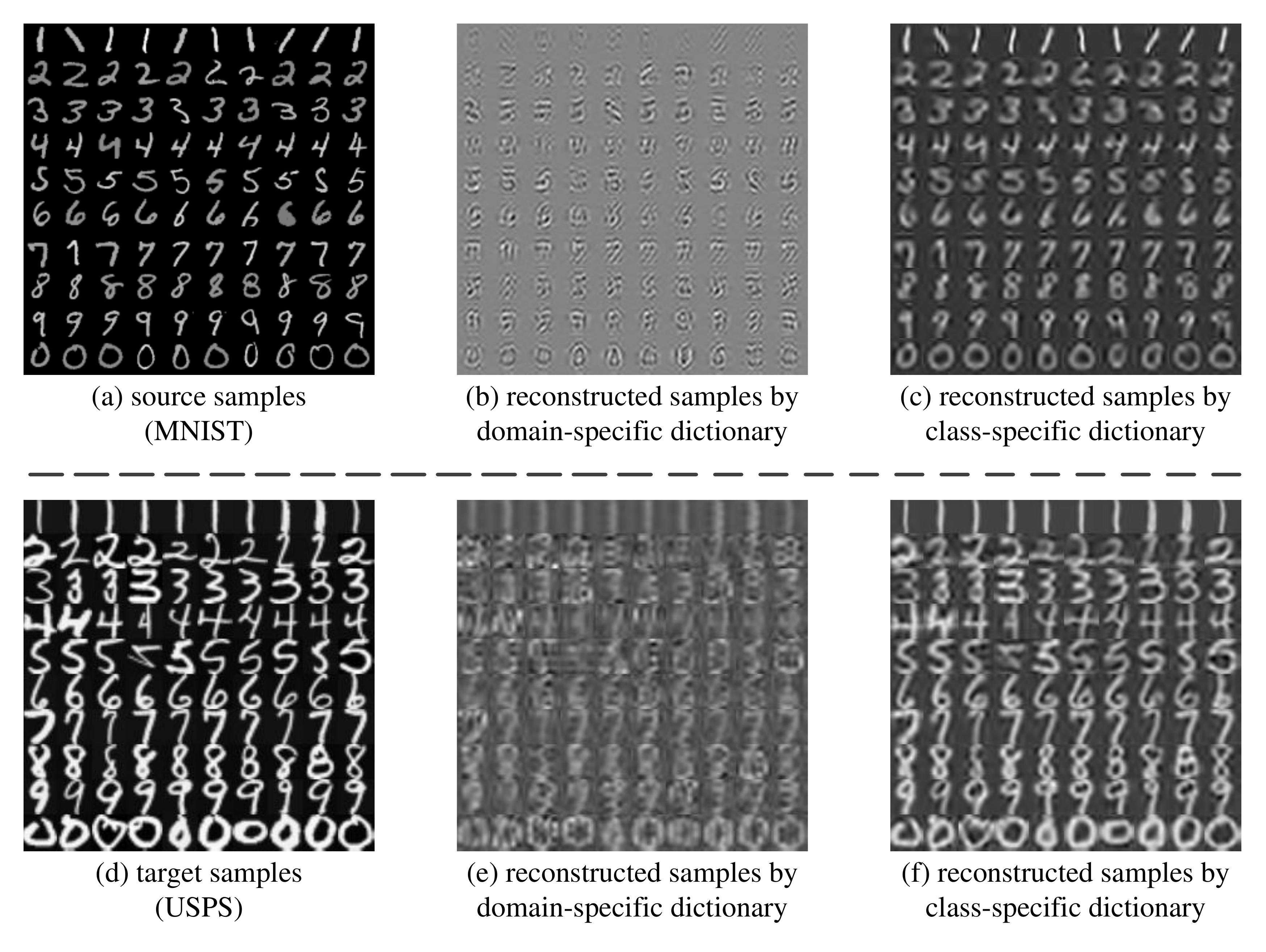}
	\caption{Visualization of reconstructed samples by domain-specific dictionary and class-specific dictionary respectively in the task MNIST $\rightarrow$ USPS. (refer to Sec.~5 for more discussion.)}
	\label{fig:reconstruction}
	\vspace{-0.4cm}
\end{figure}

To address the aforementioned issues, we propose a Structured  Discriminative Tensor Dictionary Learning (SDTDL) approach for unsupervised domain adaptation. SDTDL seeks data representation that is discrimiantive and robust to domain-shift by separating the domain factor and class factor in tensor space (Fig.~\ref{fig:reconstruction}). Specifically, a sample is factorized into domain part and class part characterized by domain-specific sub-dictionary and class-specific sub-dictionary, respectively. The resulted representation is a block diagonal sparse tensor with its nonzero blocks consisting of domain-specific representation and class-specific representation. Classification is accomplished base on reconstruction error associated with class-specific representation.

Overall, our main contributions are threefold: (1) we propose a discriminative dictionary learning approach based on tensor model for UDA. The method preserves the internal structure information of data and is able to tackle the small-sample-size problem. (2) we model domain factor and class factor separately to build a structured dictionary to guarantee the discriminativeness and domain invariance of feature. (3) exhaustive experiments on object recognition and digit recognition tasks demonstrate that the proposed SDTDL outperforms the existing shallow methods and achieves competitive results compared with deep learning approaches.

%---------------------Related Work ----------------------------
\vspace{-0.3cm}
\section{Related Work}
\vspace{-0.1cm}
Feature adaptation methods based on shallow learning include feature augmentation, feature alignment and feature transformation. \cite{Gopalan_ICCV11} and  GFK \cite{BGong_CVPR12} are two representative feature augmentation methods using intermediate subspaces to model domain shift. Subspace Alignment (SA) \cite{BFernando_ICCV13} extract linear features by aligning the subspaces of source and target domains. The feature alignment idea is extended in CORAL \cite{BSun_AAAI16} through covariance recoloring. Feature transformation methods seek a common latent feature space in which source samples and target samples are indistinguishable. The features can be obtained by linear projection \cite{Baktashmotlagh_ICCV13,MLong_CVPR14} or nonlinear mapping \cite{RAljundi_CVPR15}. %\cite{SHerath_CVPR17} tries to simultaneously reduce domain shift and preserve discriminative information in a latent Hilbert space.
%\cite{JZhang_CVPR17} considers both the statistical and geometrical characteristics of cross-domain samples to enhance the robustness of feature to domain divergence. %DsGsDL \cite{BYang_AAAI18} learns a group-sparse dictionary shared by source and target domain.
Most recently, TAISL \cite{HLu_ICCV17} is proposed to learn a tensor-form feature via Turker tensor decomposition, which is the most related work with our method. In contrast, the proposed SDTDL is able to use the valuable label information in source samples and do not need to train a classifier, which promotes the performance and efficiency in UDA.

Recently, deep convolutional neural network (CNN) based methods are developed with promising performance. Domain Adaptation Nural Network (DANN) \cite{YGanin_ICML15} combines CNN and adversarial learning to achieve an end-to-end unsupervised domain adaptation. DDC \cite{ETzeng_CVPR17} learns two feature extractors for the source and target domains respectively with GAN. DIFA \cite{RVolpi_CVPR18} extends the feature augmentation principle to generative adversarial networks. As deep UDA methods requires a large number of samples for parameter training, their effects are prone to be limited in the scenario of small sample size. By comparison, the proposed SDTDL is more suitable to address small sample size problem in domain adaptation, which is demonstrated by the experimental results in Sec.~5.

%Among UDA methods, it have been showed to be effective to assign pseudo
%labels for target samples and then add them into the training set. \cite{BLu_BMVC15} seeks a domain-robust dictionary with target label prediction in a increasing learning manner. \cite{KSaito_ICML17} assigns pseudo-labels to unlabeled target samples in the process of asymmetric tri-training for network. In \cite{RXu_CVPR18}, the classifier and the feature extractor are updated with pseudo-labeled target samples and source samples. Our method is also under the intention that discriminative knowledge transfer and accurate target label prediction can reinforce each other.

% \vspace{-0.3cm}
\section{Notations and Background}
\label{sec:notation}
% \vspace{-0.1cm}

\begin{table}[tp] \small
    \setlength{\abovecaptionskip}{-5pt}
	\caption{List of main notations.}
	\label{tab:notation}
	\begin{center}
    %\resizebox{0.1\textwidth}{!}{
        \scalebox{0.7}{
		\begin{tabular}{|l|l|l|l|}
			\hline
			Symbol                       & Description
          & Symbol                       & Description                  \\
			\hline \hline
			$\mathcal{X}$, $\mathcal{Y}$ & Tensor samples
          & $l^s, l^t$            & Class labels          \\
			\hline
			$\mathbf{U}$, $\mathbf{W}$ & Tensor dictionaries
          & $U^{(m)}$, $W^{(m)}$ & Factor matrices                  \\
            \hline			
			$\mathcal{A}$, $\mathcal{B}$ & Sparse coefficients
		  & $A_{(m)}$ & Mode-$m$ flatting of $\mathcal{A}$\\
			\hline
			$[\mathcal{A} | \mathcal{B}]$ & The stack of $\mathcal{A}$ and $\mathcal{B}$
		  &	 $\llbracket \mathcal{G};\mathbf{U} \rrbracket$ & Product of $\mathcal{G}$ with $\mathbf{U}$ \\
			\hline
		  {$\rm I$} & Identity matrix
          & $\mathbf{1}$ & Vector with all ones \\
			\hline								
		\end{tabular}}
	\end{center}
	\vspace{-0.5cm}
\end{table}

%In this section we introduce the main notations and briefly revisit the necessary preliminary knowledge on tensor algebra \cite{Kolda_SIAMReview09}.
% The basic symbols that will be used throughout the paper is listed in Table \ref{tab:notation}.
\noindent \textbf{Tensor Preliminaries.} Table \ref{tab:notation} lists the symbols  used in this paper.
An $M$-th order tensor
$\mathcal{A} \in \mathbb{R}^{I_1 \times I_2 \times \ldots \times I_M}$
is an $M$-dimensional data array, with element denoted as $a_{i_1,\ldots,i_M}$.
%The inner product of two tensors with identical size is defined as $<\mathcal{A}, \mathcal{B}> = \sum_{i_1=1,\ldots,i_m=1}^{i_1=I1,\ldots,i_M=I_M}a_{i_1,\ldots,i_M}b_{i_1,\ldots,i_M}$.
The Frobenius squared norm of $\mathcal{A}$ is defined as
$\|\mathcal{A}\|_F^2= \sum_{i_1=1,\ldots,i_m=1}^{i_1=I_1,\ldots,i_M=I_M}a^{2}_{i_1,\ldots,i_M}$. The mode-$m$ flatting of $\mathcal{A}$ reorders its elements into a matrix $A_{(m)} \in \mathbb{R}^{I_m \times \Pi_{k=1,k \neq m}^{M}I_{k}}$. The $m$-mode product of a tensor $\mathcal{A}$ with a matrix $U \in \mathbb{R}^{J_{m}\times I_{m}}$, denoted as $\mathcal{B} = \mathcal{A} \times_{m} U$, performs matrix multiplication along the $m$-th mode, which can be performed equivalently by matrix multiplication $B_{(m)} = UA_{(m)}$ and re-tensorization of undoing the mode-$m$ flattening. For conciseness and clarity, we denote the product of a tensor $\mathcal{A}$ with a set of matrices $\mathbf{U}{=}\{U^{(m)} {|} U^{(m)}{=}[u_{1}^{(m)},\ldots,u_{J_m}^{(m)}],m{=}1,\ldots,M\}$ by
\begin{equation}
	\label{eq:tensor_product}
	\llbracket \mathcal{A}; \mathbf{U} \rrbracket = \mathcal{A}
	\times_{1}U^{(1)}\times_{2} U^{(2)}\times_{3}\ldots \times_{M}U^{(M)},
\end{equation}
Similarly, We define
% \begin{equation}
%     \llbracket \mathcal{A}; \mathbf{U}_{-m}) \rrbracket{=}\mathcal{A} \times_{1}U^{(1)}\times_{2} \ldots
% 	\times_{m-1}U^{(m-1)}\times_{m+1}U^{(m+1)} \times_{m+2} \ldots
% 	\times_{M}U^{(M)}
% \end{equation}
$\llbracket \mathcal{A}; \mathbf{U}_{-m}) \rrbracket{=}\mathcal{A} \times_{1}U^{(1)}\times_{2} \ldots
	\times_{m-1}U^{(m-1)}\times_{m+1}U^{(m+1)} \times_{m+2} \ldots
	\times_{M}U^{(M)}$.

The Tucker decomposition of tensor is defined as
\begin{equation}
	\label{eq:TuckerDeco}
	\mathcal{A} = \llbracket \mathcal{G};\mathbf{U}\rrbracket
				= \sum\limits_{i_1,\dots,i_M}{g_{i_1,\dots,i_M}\mathcal{U}_{i_1,\dots,i_M}},
\end{equation}
where $g_{i_1,\dots,i_M}$ is a scale, and
$\mathcal{U}_{i_1,\dots,i_M}$
is a rank-one tensor produced by the outer product of factor vectors. Given $\mathbf{U}$, the core tensor can be obtained as
$\mathcal{G} =\llbracket \mathcal{A};\mathbf{U}^T\rrbracket$, where
$\mathbf{U}^T= \{U^{(m)T}\}_{m=1}^{M}$.
Tucker decomposition can be written in matrix format as
\begin{equation}
	\label{eq:TuckerDeco_Matrix}
	A_{(m)} = U^{(m)}H_{(m)}, \;\;\; m=1,\ldots, M,
\end{equation}
where $\mathcal{H} = \llbracket \mathcal{G};\mathbf{U}_{-m}\rrbracket$.
Note that the factor matrix $U^{(m)}$ in each mode satisfies the constraint $U^{(m)T}U^{(m)}={\rm I}$.

\noindent\textbf{Problem Definition.}
A domain $\mathscr{D}$
is composed of a feature space $\mathscr{X}$ with a marginal probability
distribution $P(\mathcal{X})$, where $\mathcal{X} \in \mathscr{X}$. A task $%
	\mathscr{T}$ associated with a specific domain $\mathscr{D}$ is defined by a
label space $\mathscr{L}$ and the conditional probability distribution $P(%
	\mathcal{L}|\mathcal{X})$, where $\mathcal{L} \in \mathscr{L}$. Domain
adaptation considers a source domain $\mathscr{D}^{s}$ and a target domain $%
	\mathscr{D}^{t}$ satisfying $\mathscr{L}^{s}=\mathscr{L}^{t}$, $\mathscr{X}^{s}=\mathscr{X}^{t}$ and $P_s(\mathcal{X}^s)\neq P_t(\mathcal{(}X)^t)$.
	%Assuming that the source domain is related to the target domain, unsupervised domain adaptation exploits the related information from $\{\mathscr{D}^s, \mathscr{T}^s\}$ to estimate $P(\mathcal{Y}|\mathcal{X}_t)$.

In this paper, we are given a set of labeled source samples
$\{\mathcal{X}_i, l_i^s\}_{i=1}^{N_s}$,
where $\mathcal{X}_{i} \in\mathbb{R}^{I_1 \times I_2 \times \ldots \times I_M}$ is a $M$th-mode tensor and $l_{i}^s \in \{1,2,\ldots, C\}$ is its class label. We are also given a set of unlabeled target samples $\{\mathcal{Y}_{j}\}_{j=1}^{N_t}$. We aim to infer the class label $l_j^{t}\in \{1,2,\ldots, C\}$ of $\mathcal{Y}_{j}$ by learning from the source and target samples.

%======section: Structural Discriminative Tensor Dictionary Learning ===
% \vspace{-0.1cm}
\section{The Proposed SDTDL}
% \vspace{-0.1cm}
\subsection{Formalization}
% \vspace{-0.1cm}
For easy understanding, we assume the labels of target samples have been predicted at present, and provide the details of label prediction and target sample selection in sec \ref{subsec:ple}. We select partial target samples based on their prediction confidence to be additional training samples to aid modal training.
The selected $N_{t}^c$ target samples from the $c$-th class are denoted as $\mathcal{Y}^{c} \in \mathbb{R}^{I_1 \times I_2 \times \ldots \times I_M \times N_{t}^c}$, and the $N_{s}^c$ source samples belonging to the $c$-th class are denoted as $\mathcal{X}^{c}\in \mathbb{R}^{I_1 \times I_2 \times \ldots \times I_M \times N_{s}^c}$.

We model the generation process of cross-domain data as the combination of domain factor and class factor, in which a sample $\mathcal{Z}$ %($\mathcal{Z}=\mathcal{X} or \mathcal{Y}$)
($\mathcal{Z} \in \{\mathcal{X}, \mathcal{Y}\}$)  is factorized as
%*******************************************************************************
\begin{equation}
	\label{eq:factorization}
	\mathcal{Z} = \mathcal{Z}_{domain} + \mathcal{Z}_{class},
\end{equation}
%*******************************************************************************
where $\mathcal{Z}_{domain}$ is determined by the unique character of the domain from which $\mathcal{Z}$ is sampled and $\mathcal{Z}_{class}$ is determined by the semantic information of the class to which $\mathcal{Z}$ belongs.

%From the perspective of representation theory, formula (\ref{eq:TuckerDeco}) provides a ``dense'' representation of tensor $\mathcal{A}$ with $\mathcal{U}_{i_1,\dots,i_M}$ as a atom of tensor dictionary and $g_{i_1,\dots,i_M}$ as the corresponding representation coefficient over the dictionary.

In order to obtain `` parsimonious'' representations of $M$-th order tensor samples, we propose to learn a structured tensor dictionary $\mathbf{D}$ composed of $M$ factor matrices, i.e. $\mathbf{D}{=}\{D^{(m)}\}_{m= 1}^{M}$. The structure of $\mathbf{D}$ arises from the structure of each factor matrix. Specifically, $D^{(m)}$ is composed of domain-specific sub-dictionary $U^{(m)}$ and class-shared sub-dictionary matrix $W^{(m)}$, i.e. $D^{(m)} = [U^{(m)},W^{(m)}]$. In order to distinguish source domain from target domain, $U^{(m)}$ is further divided into source-specific sub-dictionary $U_{s}^{(m)}$ and target-specific sub-dictionary $U_{t}^{(m)}$. This leads us to the following factorization of a source sample
%*******************************************************************************
\begin{equation}
	\label{eq:domain_dict_rep}
	\mathcal{X}_{domain} = \llbracket \mathcal{A}_{0}; \mathbf{U}_{s} \rrbracket,
\end{equation}
%*******************************************************************************
where $\mathbf{U}_{s}{=}\{U_{s}^{(m)}\}_{m= 1}^{M}$ is the source-specific sub-dictionary, and $\mathcal{A}_{0}$ is the domain representation of $\mathcal{X}$ in tensor format. Similarly, we have $\mathcal{Y}_{domain} = \llbracket \mathcal{B}_{0}; \mathbf{U}_{t}  \rrbracket$ for target sample $\mathcal{Y}$ with the target-specific sub-dictionary $\mathbf{U}_{t}$.
%Note that the domain-specific representation of a sample over $\mathbf{U}$ is actually a sparse block diagonal tensor.

Model (\ref{eq:factorization}) indicates that $\mathcal{Z}_{class}$ is merely determined by class factor, thus it's safe to assume that $\mathcal{X}_{class}$ and $\mathcal{Y}_{class}$ can be represented over a shared sub-dictionary $\mathbf{W}$. Due to the success of structured discriminative dictionary learning in image classification \cite{MYang_ICCV11}, we divide $\mathbf{W}$ into a serial of class-specific sub-dictionaries $\mathbf{W}_1, \ldots, \mathbf{W}_{C}$ for discriminative representation. %The class-specific subdictionaries should be non-overlapping in order to provide structural sparse representation for different classes, therefore the class-specific factor dictionary matrix can be expressed as $W^{(m)} = [W^{(m)}_1,\ldots, W^{(m)}_C]$.
To $\mathcal{X}$ from class $c$, its tensor representation over $\mathbf{W}_{c}$ is given by
%*******************************************************************************
\begin{equation}
	\label{eq:discrim_rep}
	\mathcal{X}_{class} = \llbracket \mathcal{A}_{c}; \mathbf{W}_{c} %
	\rrbracket,
\end{equation}
%*******************************************************************************
where $\mathcal{A}_{c}$ is the class-specific representation. Similarly, we have
$\mathcal{Y}_{class} = \llbracket \mathcal{B}_{c}; \mathbf{W}_{c} \rrbracket$, where $\mathcal{B}_{c}$ provides the class-specific representation.

Based on our notation of $\mathcal{X}^{c}$ and $\mathcal{Y}^{c}$ in section \ref{sec:notation}, we define
$\mathcal{A}_{c}^{c} \in \mathbb{R}^{J_1 \times J_2 \times \ldots \times J_M \times N_{s}^{c}}$ and $\mathcal{B}_{c}^{c} \in \mathbb{R}^{J_1 \times J_2 \times \ldots \times J_M \times N_{t}^{c}}$ as the representations of source sample set $\mathcal{X}^{c}$ and target sample set $\mathcal{Y}^{c}$ over the shared $\mathbf{W}_{c}$ of class $c$ respectively. In order to correct the domain shift, the class-conditional distributions of representation in source and target should be aligned. Here we adopt Maximum Mean Discrepancy \cite{AGretton_JMLR12} to measure distribution divergence, then we have
%*******************************************************************************
\begin{equation}
	\label{eq:MMD}
	\|\mathcal{M}_{c}^{\mathcal{A}} - \mathcal{M}_{c}^{\mathcal{B}} \|_{F}^{2},
\end{equation}
%*******************************************************************************
where $\mathcal{M}_{c}^{\mathcal{A}} = \frac{1}{N_s^{c}}\sum_{i}{\mathcal{A}_{c,i}^c}$
and $\mathcal{M}_{c}^{\mathcal{B}} = \frac{1}{N_t^{c}}\sum_{j}{\mathcal{A}_{c,j}^c}$. Beyond that, the intra-class variance of representation should be small to facilitate the discriminativeness. To that end, the following objective is to be minimized for source domain
%*******************************************************************************
\begin{equation}
	\label{eq:intra-class}
	\|\mathcal{A}_{c}^{c} - \mathcal{\bar{M}}_{c}^{\mathcal{A}} \|_{F}^{2},
\end{equation}
%*******************************************************************************
where $\mathcal{\bar{M}}_{c}^{\mathcal{A}}$ is produced by arranging $N_s^c$ duplicate $\mathcal{M}_{c}^{\mathcal{A}}$ so that $\mathcal{\bar{M}}_{c}^{\mathcal{A}}$ and $\mathcal{A}_{c}^{c}$ have the same size. In the same way,
$\|\mathcal{B}_{c}^{c} - \mathcal{\bar{M}}_{c}^{\mathcal{B}} \|_{F}^{2}$ should be minimized for target domain. To satisfy both (\ref{eq:MMD}) and (\ref{eq:intra-class}), we need to minimize the following objective
%*******************************************************************************
\begin{equation}
	\label{eq:reqularization}
	\|\mathcal{A}_{c}^{c} - \mathcal{\bar{M}}_{c}^{B}\|_{F}^2 + \|\mathcal{B}_{c}^{c}
	- \mathcal{\bar{M}}_{c}^{A}\|_{F}^2.
\end{equation}
%*******************************************************************************

By considering the above criteria together, our learning model can be written as
%*******************************************************************************
\begin{equation}
	\label{eq:model}
	\begin{aligned}
		\underset{_{(\mathcal{A}, \mathcal{B}, \mathbf{U},
		\mathbf{W})}}{\textmd{arg min}} & \; \sum_{c=1}^{C} \left \{
		\begin{aligned}
			  & \;\; \|\mathcal{X}^{c} - \llbracket \mathcal{A}_{0}^{c};\mathbf{U}_{s} \rrbracket
			- \llbracket  \mathcal{A}_{c}^{c};\mathbf{W}_{c} \rrbracket \|_{F}^{2} \\
			+ & \theta \|\mathcal{Y}^{c} - \llbracket \mathcal{B}_{0}^{c}; \mathbf{U}_{t}\rrbracket
			- \llbracket \mathcal{B}_{c}^{c}; \mathbf{W}_{c} \rrbracket \|_{F}^{2}  \\
			+ & \lambda \left (\|\mathcal{A}_{c}^{c} - \mathcal{\bar{M}}_{c}^{B}\|_{F}^2
			+ \|\mathcal{B}_{c}^{c} - \mathcal{\bar{M}}_{c}^{A}\|_{F}^2 \right ) \\
		\end{aligned}
		\right \} \\
		\text{s.\;t.}  & \;\;\; U_s^{(m)T}U_s^{(m)}={\rm I}, \;\; U_t^{(m)T} U_t^{(m)}=I, \\
				& \;\;\; W_{c}^{(m)T} W_{c}^{(m)} = {\rm I}, \;\;\;m=1,2,\ldots,M.	
	\end{aligned}
\end{equation}
%*******************************************************************************
The first and second terms are the fidelity of the reconstruction over the structured tensor dictionary. The third term can be viewed as discriminant analysis of the representation. $\theta$ determines the weighting of target domain compared with source domain, and $\lambda$ trades off between fidelity term and discriminative term. The constraints require that the factor matrices in each mode are orthogonal matrices.

% %By minimizing the objective function, we jointly learn the domain-specific
% dictionary to describe the domain information and the class-specific dictionary
% to provide discriminative feature. It should be noticed that the overall representation of a sample over tensor dictionary $\mathbf{D}$ is a block diagonal tensor with highly structural sparse characteristic.

%==================FLAG ---> Subsection: Optimization ==================
% \vspace{-0.1cm}
\subsection{Optimization}
% \vspace{-0.1cm}
In this section, we solve model (\ref{eq:model}) using alternative optimization strategy, in which %the variables are divided into three groups: $\mathbf{U}_s$ and $\mathcal{A}_{0}$, $\mathbf{U}_{t}$ and $\mathcal{B}_{0}$, $\mathbf{W}_c$,$\mathcal{A}_c$ and $\mathcal{B}_{c}$.
we seek the optimal solution for some certain variables while keeping all the others fixed at the values of the previous iteration till the iteration converges. %The convergence property of our method is experiencel evaluated in section, which show our method convergence always convergence with $20$ iterations.

% \subsubsection{Alternative Optimization}
\noindent \textbf{Optimize $\{\mathcal{A}_0,\mathbf{U}_{s}\}$.}
With the fixed $\{\mathbf{W}_{c}, \mathcal{A}_{c}^{c}\}_{c=1}^{C}$, the fidelity loss in regard to class $c$ can be written as
$\mathcal{\bar{X}}^{c}- \llbracket \mathcal{A}_{0};\mathbf{U}_{s} \rrbracket$, where $\mathcal{\bar{X}}^{c}=\mathcal{X}^{c} - \llbracket  \mathcal{A}_{c}^{c};\mathbf{W}_{c} \rrbracket$. Considering all the source samples, model (\ref{eq:model}) becomes
%****************************************
\begin{equation}
	\label{eq:submodel_A0}
	\begin{aligned}
		\underset{_{\mathcal{A}_{0}, \mathbf{U}_{s}}}
		{\textmd{arg min}} & \;
		\|\mathcal{\bar{X}} - \llbracket \mathcal{A}_{0};\mathbf{U}_{s} \rrbracket \|_{F}^{2} \\
		\text{s.\;t.}  & \;\;\; U_s^{(m)T}U_s^{(m)}={\rm I}, \;\;\;m=1,2,\ldots,M.	
	\end{aligned}
\end{equation}
%****************************************
Model (\ref{eq:submodel_A0}) is a typical best rank-$(J_1,J_2,\ldots,J_M)$ tensor approximation problem that can be solved by HOOI algorithm \cite{LLathauwer_SIAM_JMAA00}.

\noindent\textbf{Optimize $\{\mathcal{B}_0,\mathbf{U}_{t}\}$.}
In the same way as in (\ref{eq:submodel_A0}), the target-specific dictionary and the domain-specific representation of target samples are obtained by applying HOOI to the following optimal problem.
%****************************************
\begin{equation}  \label{eq:submodel_B0Ut}
	\begin{aligned}
		\underset{_{\mathcal{B}_{0}, \mathbf{U}_{t}}}
		{\textmd{arg min}} & \;
		\|\mathcal{\bar{Y}} - \llbracket \mathcal{B}_{0};\mathbf{U}_{t} \rrbracket \|_{F}^{2} \\
		\text{s.\;t.}  & \;\;\; U_t^{(m)T}U_t^{(m)}={\rm I}, \;\;\;m=1,2,\ldots,M,	
	\end{aligned}
\end{equation}
%****************************************
where $\mathcal{\bar{Y}}^{c}=\mathcal{Y}^{c} - \llbracket  \mathcal{B}_{c}^{c};\mathbf{W}_{c} \rrbracket$.

\noindent\textbf{Optimize $\{\mathcal{A}_{c}^{c},\mathcal{B}_{c}^{c},\mathbf{W}_{c}\}$.}
%As there is no overlap between the class-wise subdictionaries in $\mathbf{W}$,
We seek the optimal sub-dictionary class-by-class, so we have the model for class $c$ as
%****************************************
\begin{equation}  \label{eq:submodel_AcWc}
	\begin{aligned}
		\underset{_{\mathcal{A}_{c}^{c},\mathcal{B}_{c}^{c},\mathbf{W}_{c}}}
		{\textmd{arg min}} & \; \left \{
		\begin{aligned}
			  & \|\mathcal{\widetilde{X}}^{c} - \llbracket \mathcal{A}_{c}^{c};\mathbf{W}_{c} \rrbracket \|_{F}^{2}
			+ \theta \|\mathcal{\widetilde{Y}}^{c} - \llbracket \mathcal{B}_{c}^{c};\mathbf{W}_{c} \rrbracket \|_{F}^{2} \\
			+ & \lambda \left ( \|\mathcal{A}_{c}^{c} - \mathcal{\bar{M}}_{c}^{B} \|_{F}^{2}
			+ \|\mathcal{B}_{c}^{c} - \mathcal{\bar{M}}_{c}^{A} \|_{F}^{2} \right ) \\
		\end{aligned}
		\right \} \\
		% s.\;t.             & \; \mathcal{A}_{c}^{c} \in R^{J_1\times J_2 \times \ldots \times J_M},
		% \; \mathcal{B}_{c}^{c} \in R^{J_1\times J_2 \times \ldots \times J_M}, \\
		\text{s.\;t.}  & \; W_c^{(m)T}W_c^{(m)}={\rm I}, \quad m=1,\ldots,M,	
	\end{aligned}
\end{equation}
%****************************************
where $\mathcal{\widetilde{X}}^{c} = \mathcal{X}^{c} - \llbracket \mathcal{A}_{0}^{c};\mathbf{U}_{s} \rrbracket$, $\mathcal{\widetilde{Y}}^{c} = \mathcal{Y}^{c} - \llbracket \mathcal{B}_{0}^{c};\mathbf{U}_{t} \rrbracket$.

%The key point to solve model (\ref{eq:submodel_AcWc}) is to seek the optimal factor matrices in each mode that compose the tensor dictionary.
We adopt the alternating optimization strategy in \cite{LLathauwer_SIAM_JMAA00} to update $\mathbf{W}_{c}$ and $\{\mathcal{A}_{0}^c, \mathcal{B}_{0}^c \}$ by turns. With fixed $\mathcal{A}_{c}^c$ and  $\mathcal{B}_{c}^c$, the optimal $\mathbf{W}_{c}$ is provided by theorem \ref{th:th1}. With fixed $\mathbf{W}_{c}$, the optimal $\{\mathcal{A}_c^c, \mathcal{B}_{c}^c \}$  is given by $\mathcal{A}_c^c =\llbracket \mathcal{\widetilde{X}}^{c};\mathbf{W}_{c}^T\rrbracket$ and
$\mathcal{B}_c^c =\llbracket \mathcal{\widetilde{Y}}^{c};\mathbf{W}_{c}^T\rrbracket$.

\begin{theorem}
\label{th:th1}
Let $\mathcal{\widetilde{Z}}^{c}=[\mathcal{\widetilde{X}}^{c}|\mathcal{\widetilde{Y}}^{c}]$ be the augmented sample tensor of class $c$ which is generated by concatenating $\mathcal{\widetilde{X}}^{c}$ and $\mathcal{\widetilde{Y}}^{c}$ along with the $M+1$-th mode. %Similarly, denote by $\mathcal{G}^{c}=[\mathcal{A}^{c}_c | \mathcal{B}^{c}_c]$ the augmentation of represenation.
Define matrix $\Phi$ as
%****************************************
\begin{equation}
	\Phi = \left(
	\begin{array}{cc}
			(1-\sqrt{\lambda}){\rm I}_{N_s^c}                    & \frac{\sqrt{\lambda}}{N_s^c}1_{N_s^c}1^{T}_{N_t^c} \\
			\frac{\sqrt{\lambda}}{N_t^c}1_{N_t^c}1^{T}_{N_s^c} & (\sqrt{\theta}-\sqrt{\lambda}){\rm I}_{N_t^c}
		\end{array}\right),
\label{eq:def_Phi}
\end{equation}
%****************************************
where ${\rm I}_{N_s^c}$ and ${\rm I}_{N_t^c}$ are identical matrices, $\mathbf{1}_{N_s^c}$ and $\mathbf{1}_{N_t^c}$ are the column vectors with all ones. Let $G_c^{(m)}$ be the mode-$m$ flatting matrix of $\mathcal{G}_{c} = \llbracket \mathcal{\widetilde{Z}}^{c};W^{(1)T}_c,\ldots, W^{(m-1)T}_c, {\rm I}, W^{(m+1)T}_c,\ldots,W^{(M)T}_c, \Phi \rrbracket$.
Then, the optimal $\mathbf{W}_c$ in (\ref{eq:submodel_AcWc}) is provided by $W_{c}^{(m)}=[\eta_{1}^{(m)},\ldots,\eta_{J_m}^{(m)}]$, with columns as the eigen-vectors corresponding to the first largest $J_m$ eigenvalues of the following eigenvalue-problem
%****************************************
\begin{equation}
% 	\widetilde{G}^{c,m}_{(m)}\big(\widetilde{G}^{c,m}_{(m)}\big)^T\eta = \alpha\eta
G_c^{(m)}\big(G_c^{(m)}\big)^T\eta = \alpha\eta.
	\label{eq:eigen-value}
\end{equation}
%****************************************
\end{theorem}

\noindent The proof is given in Appendix A.

\subsection{Label Prediction and Sample Selection}
% \vspace{-0.1cm}
\label{subsec:ple}
%In model (\ref{eq:model}), the class labels of target samples are required in advance.
%In this section, the target labels are predicted using posterior probability so to avoid classifier design and training. Concretely,
The probability of $\mathcal{Y}_{j}$ belonging to class $c$ can be computed based on the fidelity error, i.e.
%****************************************
\begin{equation}
	P_{c|j}^{r} = \frac{\text{exp}(-\|\mathcal{Y}_{j} - \llbracket \mathcal{B}_{0}^{j}; \mathbf{U}_{t}\rrbracket
	- \llbracket \mathcal{B}_{c}^{j}; \mathbf{W}_{c} \rrbracket \|_{F}^{2}/\sigma)}{\sum_{k}{\text{exp}(-\|\mathcal{Y}_{j} - \llbracket \mathcal{B}_{0}^{j}; \mathbf{U}_{t}\rrbracket
	- \llbracket \mathcal{B}_{k}^{j}; \mathbf{W}_{k} \rrbracket \|_{F}^{2}/\sigma)}},
	\label{eq:pb_err}
\end{equation}
%****************************************
where $\sigma$ is the parameter of exponent function whose value is set as median value of the denominator.
%we compute the reconstruction error based on domain-wise and class-specific dictionary as follows
%****************************************
% \begin{equation}  \label{eq:eigen-value}
% 	\mathcal{E}^{c}_r = \|\mathcal{Y}^{c} - \llbracket \mathcal{B}_{0}^{c}; \mathcal{U}_{t}\rrbracket
% 	- \llbracket \mathcal{B}_{c}^{c}; \mathcal{W}_{c} \rrbracket \|_{F}^{2}
% \end{equation}
%****************************************
The posterior probability can also be computed based on the deviation of $\mathcal{Y}_{j}$ from the centroid of class $c$. Thus we have
%****************************************
\begin{equation}
	P_{c|j}^{d} = \frac{\text{exp}(-\|\mathcal{B}_{j}^{c} - \mathcal{M}_{c}^{\mathcal{A}}\|_{F}^2/\sigma)}{\sum_{k}{\text{exp}(-\|\mathcal{B}_{j}^{c} - \mathcal{M}_{k}^{\mathcal{A}}\|_{F}^2/\sigma)}}.
	\label{eq:pb_dev}
\end{equation}
%****************************************
$\mathcal{M}_{c}^{\mathcal{A}}$ is adopted to replace $\mathcal{M}_{c}^{\mathcal{B}}$ for two reasons: (1) $\mathcal{M}_{c}^{\mathcal{A}}$ is more reliable because it is computed according to the real source labels; (2) it is beneficial to alleviate the domain shift to make the target sample towards to the corresponding class center in source domain.

Through the convex combination of the two kinds of probabilities, We ultimately can predict the class label of $\mathcal{Y}_{j}$ by
%****************************************
\begin{equation}
	l_{j}^t = \textmd{arg} \;\underset{_{c}}
	{\textmd{min}} \;\; P_{c|j} = \textmd{arg} \;\underset{_{c}}
	{\textmd{min}} \;\; \gamma P_{c|j}^r + (1-\gamma) P_{c|j}^d.
\label{eq:predictlabel}
\end{equation}
%****************************************

In order to select target samples with reliable pseudo-labels for training, we sort $P_{c|j}$s in descend order. Then we add the target samples associated with highest poster probability into training sample set. The ratio $\delta$ of the selected target samples in the whole target sample set is a parameter of our model.

%--------------------The PyseudoCode of the algorithm -----------------------
\begin{algorithm}[!t]\small
	\caption{The Proposed SDTDL.}
	\label{alg:SDTDL}
	\begin{algorithmic}[1]
	  \Require
		The labeled source samples $\{\mathcal{X}_i, l_i^s\}_{i=1}^{N_s}$;
		target samples $\{\mathcal{Y}_i\}_{i=1}^{N_t}$;
		parameters:$\theta$,$\lambda$, $\gamma$, $\delta$, feature dimension in each mode $\{J_i\}_{i=1}^{M}$;
	  \Ensure
		The class label of target samples $\{l_j^t\}_{j=1}^{N_t}$; source-specific tensor dictionary~$\mathbf{U}_s$; target-specific  dictionary~$\mathbf{U}_t$; class-specific dictionary $\{\mathbf{W}_c\}_{c=1}^{C}$.
	  \State Initialize $\{\mathbf{W}_c\}_{c=1}^{C}$, $\mathbf{U}_s$,  $\mathbf{U}_t$ in turn.
	  \Repeat
		\State Predict target label $\{l_j^t\}_{j=1}^{N_t}$ by (\ref{eq:predictlabel});
		\State Select target samples for training model;
		\State Compute $\{W_c^{(m)}\}_{m=1}^{M}$ by (\ref{eq:eigen-value}) to update $\{\mathbf{W}_c\}_{c = 1}^{C}$;
		\State Update $\mathbf{U}_s$ by solving (\ref{eq:submodel_A0});
		\State Update $\mathbf{U}_t$ by solving (\ref{eq:submodel_B0Ut});
	  \Until{stopping criteria is reached.}
	\end{algorithmic}
  \end{algorithm}
%----------------------------------------------------------------------------

%=========================FLAG Initialization =============================

%-------------------Figure: Visualization of features ---------------
\begin{figure*}[t]\small
	\centering
	\includegraphics[width=1\linewidth]{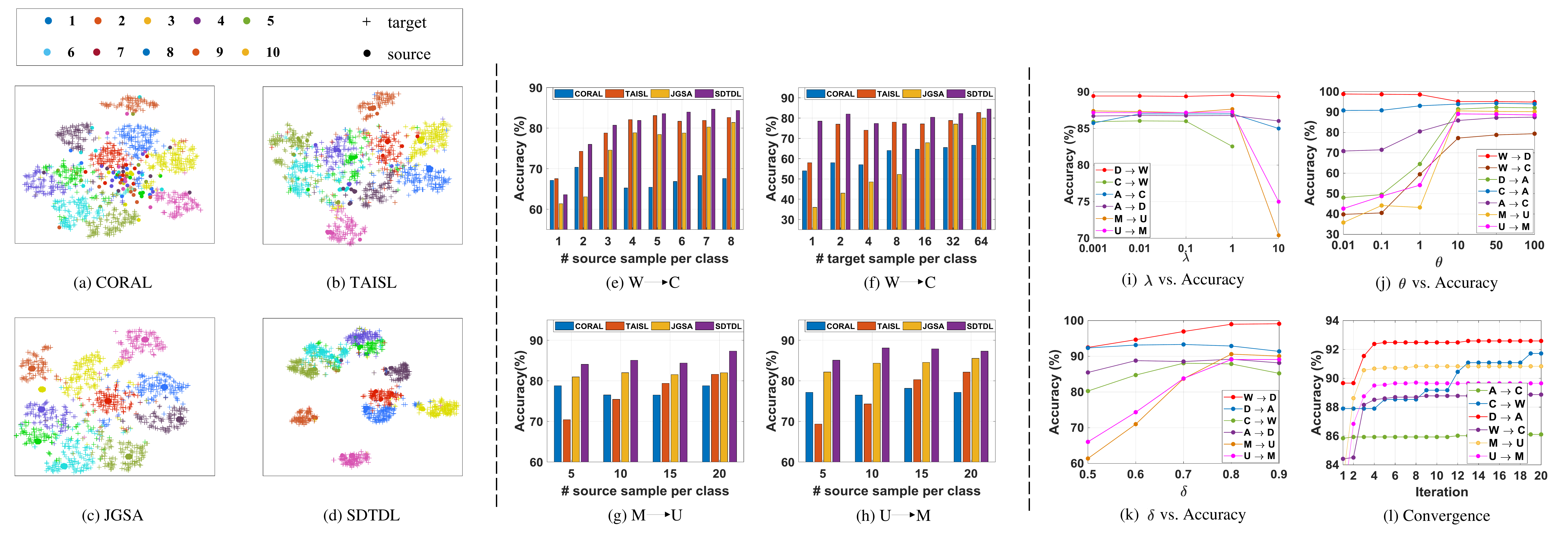}
	\caption{Visualization of feature on the DA task of W $\rightarrow$ C (a-d), accuracy rates with varying number of samples per class from source domain (e,g,h) and target domain (f), accuracy rates with varying values of parameters (i-k) and the number of iterations (l). (Better view zoomed in and with color)
% 	(Better view zoomed in and with color)
	}
	\label{fig:visualization}
	\vspace{-0.3cm}
\end{figure*}

% \vspace{-0.1cm}
\subsection{Initialization}
% \vspace{-0.1cm}
\label{subsec:initialization}
The initialization process includes the following three steps. In step 1, the class-specific dictionary $\mathbf{W}$ are initialized by structured discriminant dictionary learning (e.g. \cite{MYang_ICCV11}) based on the labeled source samples, followed by computing class-wise sparse coding $\mathcal{A}_c$ and $\mathcal{B}_c$. Then the domain-specific dictionary $\mathbf{U}_s$ are initialized through (\ref{eq:submodel_A0}).	
In step 2, the target labels are predicted by (\ref{eq:predictlabel}) without the influence of $\mathbf{U}_t$, i.e. set $\llbracket \mathcal{B}_{0}^{j}; \mathbf{U}_{t}\rrbracket{=}0$ in (\ref{eq:pb_err}). Note that at this stage, although the estimated target labels may be deviated from the actual ones,they provide a reasonable start point for iteration because of the underlying correlation between source and target.
In step 3, we select partial target samples with their estimated labels to initialize the target-specific dictionary $\mathbf{U}_t$ through (\ref{eq:submodel_B0Ut}).
In summary, the proposed method can be expressed in Algorithm~\ref{alg:SDTDL}.

%$\mathbf{
%\mathcal{X}^{c} \; \mathcal{Y}^{c} \; \\
%U^{(1)}_{s} \; U^{(2)}_{s}  \;  U^{(3)}_{s}\; \\
%U^{(1)}_{t}\; U^{(2)}_{t} \; U^{(3)}_{t} \;
%W^{(1)}_{1} \\ W^{(1)}_{c} \; W^{(1)}_{C} \;
%W^{(2)}_{1} \; W^{(2)}_{c} \; W^{(2)}_{C}\\
%W^{(3)}_{1} \; W^{(3)}_{c} \; W^{(3)}_{C} \\
%\mathcal{A}^{c}_{0} \; \mathcal{B}^{c}_{0}\;
%\mathcal{A}^{c}_{c} \; \mathcal{B}^{c}_{c}\;
%D^{(1)} \; D^{(2)} \; D^{(3)}
%}$

% \vspace{-0.1cm}
\section{Experiments}
% \vspace{-0.1cm}
%--------------------Datasets-------------------------------------------------
\subsection{Experimental Setup}
% \vspace{-0.1cm}
\noindent\textbf{Datasets.} We employ two pubic datasets to evaluate the propsoed method. (1) Office+Caltech dataset is released by \cite{BGong_CVPR12}, which consists of $2,533$ images of $10$ object classes from $4$ domains, \emph{i.e.}, Amazon (A), Webcam (W), Dslr (D) and Caltech (C). We randomly select $8$ labeled images per class from Webcam/DSLR/Caltech and $20$ from Amazon as source samples respectively according to \cite{BGong_CVPR12}. We ran $20$ different trials and report the average rate and standard deviation of recognition accuracy. For fair comparison, we use the tensor data provided by \cite{HLu_ICCV17}, which is produced by CONV5\_3 layer of the VGG-16 model. For other methods, we report the results in the literature.
(2) To evaluate the performance of the methods in the settings of small sample size, we adopt the USPS+MNIST dataset released by \cite{MLong_CVPR14}, which consists of $1,800$ digital images from USPS and $2,000$ digit images in MNIST from $0$ to $9$. Thus these two domains lead to two DA tasks.
The tensor samples are produced by CONV5\_3 layer of VGG-16 model pre-trained with all the data in MNIST.
% The details of data pre-processing is in the supplementary material.

\noindent\textbf{Baseline Models.} The proposed SDTDL is compared with seven competitive UDA methods, \emph{i.e.}, No Adaptation (NA), TCA \cite{SPan_TNN11}, GFK \cite{BGong_CVPR12}, DIP \cite{Baktashmotlagh_ICCV13}, SA \cite{BFernando_ICCV13}, LTSL \cite{MShao_IJCV14}, LSSA \cite{RAljundi_CVPR15}, and three state-of-the-art UDA methods, \emph{i.e.}, CORAL \cite{BSun_AAAI16}, TAISL \cite{HLu_ICCV17} and JGSA \cite{JZhang_CVPR17}.
For digit recognition task, two deep UDA methods DANN \cite{YGanin_ICML15} and DDC \cite{ETzeng_CVPR17} are added into comparison to evaluate SDTDL in small sample size scenario.

\noindent\textbf{Parameter Settings.}
%According to \cite{HLu_ICCV17}, One-vs-rest linear SVMs are used for recognition in the baseline methods.
The optimal parameters of SDTDL are set empirically based on grid searching. Specifically, for object recognition, the parameters are set as: $\theta{=}20$, $\lambda{=}0.1$, $\gamma{=}0.25$,
$\delta{=}0.8$, $dim{=}[6,6,28]$. For digital recognition task, the parameters are set as: $\theta{=}10$, $\lambda{=}1$, $\gamma{=}0.2$,
$\delta{=}0.8$, $dim{=}[7,7,30]$. %as shown in Table \ref{table:parameter}.
The parameters of the other methods in comparison are set according to the corresponding papers.

%\begin{table}[t]
%	\caption{Parameter settings of SDTDL.}	
%		\label{table:parameter}
%	\begin{center}		
%% 		\resizebox{0.4\textwidth}{!}{
%		\begin{tabular}{|l|l|l|l|}
%			\hline
%			\multicolumn{2}{|c|}{Object Recognition} & \multicolumn{2}{|c|}{Digit Recognition} \\
%			\hline
%			parameter  & value  &  parameter & A value \\
%			\hline
%			$\theta$  & 20 & $\theta$ & 15\\
%			\hline
%			$\lambda$  & 0.1 & $\lambda$ & 0.1\\
%			\hline
%			$\gamma$  & 0.25 & $\gamma$ & 0.1\\
%			\hline
%			$\delta$  & 0.8 & $\delta$ & 0.7\\
%			\hline
%			$dim$  & $[6,6,28]$ & $dim$ & $[7,7,60]$\\
%			\hline
%					
%		\end{tabular}
%% 		}
%	\end{center}
%\end{table}

%--------------------Task-------------------------------------------------
% \vspace{-0.1cm}
\subsection{Experimental Results}
% \vspace{-0.1cm}
\noindent\textbf{Feature Visualization.}
To qualitatively evaluate the discriminativeness and robustness to domain-shift of the feature extracted by SDTDL, we visualize the feature embeddings in the domain pair Webcam to Caltch (W$\rightarrow$C).
We compare SDTDL with CORAL, TAISL and JGSA in terms of the $2$D scatter plot given by t-distributed stochastic neighbor embedding (t-SNE) \cite{Maaten_IJML15}.
Fig.~\ref{fig:visualization}~(a-d) illustrate the visualized distributions of the features corresponding to source and target samples. The features extracted by SDTDL are more prone to form separate clusters associated with the $10$ categories compared with other baselines.
For both source and target samples, the intra-class scatter is small and the inter-class scatter is large, indicating that SDTDL is able to guarantee the feature to be discriminative. Besides, the distributions of source samples and target samples are aligned for each category, which suggests that our method can suppresses the interference of domain factor to discriminative information transfer from the source domain to the target domain.
%for CORAL, although the target features are prone to form seperate clusters, the mixed source features provide less information of category. This result is in accorance with the performance of CORAL in table \ref{tab:Accy_OC10_routine}.

%------------------------Table: Accy_OC10 --------------------------------------
\begin{table*}[t]\small
	\caption{Accuracy ($\%$) on cross-domain visual
    datasets. \textbf{Bold} numbers indicate the best results.}
    \label{tab:Accy_OC10_routine}
	\centering	
	\resizebox{\textwidth}{!}{
	\begin{tabular}{|l | c c c c c c c c c c c c |c|}
	\hline
	Method 	& C $\rightarrow$ A	& C $\rightarrow$ W & C $\rightarrow$ D
			& A $\rightarrow$ C	& A $\rightarrow$ W & A $\rightarrow$ D
			& W $\rightarrow$ C	& W $\rightarrow$ A & W $\rightarrow$ D
			& D $\rightarrow$ C	& D $\rightarrow$ A & D $\rightarrow$ W
			& MEAN \\
	\hline
	\hline
	NA 			& 89.0(2.0) & 79.4(2.7) & 86.2(4.0)
				& 77.3(1.8) & 74.6(3.1) & 82.8(2.2)
				& 63.7(2.1) & 74.0(2.5) & 94.9(2.4)
				& 70.5(1.9) & 81.1(1.9) & 91.1(1.7) & 80.4 \\
	TCA 		& 78.1(6.1) & 69.0(6.6) & 74.3(5.2)
				& 56.7(4.5) & 55.5(6.4) & 59.9(6.7)
				& 54.7(3.8) & 68.3(4.1) & 90.6(3.2)
				& 51.9(2.2) & 61.2(4.2) & 89.9(2.2) & 67.5 \\
	GFK 		& 87.6(2.3) & 81.9(4.9) & 84.8(.45)
				& 75.1(3.9) & 74.3(5.2) & 81.4(4.3)
				& 79.1(2.7) & 84.0(4.4) & 95.2(2.2)
				& 82.2(2.4) & 90.4(1.4) & 92.8(2.2) & 84.1 \\
	DIP 		& 84.8(4.3) & 73.5(4.9) & 82.8(7.7)
				& 59.8(5.7) & 45.5(9.1) & 52.2(8.1)
				& 65.2(4.5) & 69.3(6.9) & 94.1(3.1)
				& 61.9(6.3) & 76.4(3.7) & 90.9(2.3) & 71.4 \\
	SA 			& 82.0(2.6) & 65.9(4.0) & 73.7(4.3)
				& 67.7(4.2) & 61.1(5.1) & 67.8(4.8)
				& 70.4(4.1) & 80.1(4.3) & 91.1(3.3)
				& 66.9(3.3) & 77.4(6.0) & 87.3(3.1) & 74.3 \\
	LTSL 		& 87.5(2.8) & 75.3(4.2) & 82.3(4.1)
				& 70.2(2.4) & 66.7(4.6) & 77.7(4.6)
				& 59.1(4.4) & 66.6(5.7) & 90.0(3.8)
				& 60.8(3.1) & 69.2(4.5) & 86.0(2.9) & 74.3 \\
	LSSA 		& 86.4(1.7) & 45.4(6.6) & 73.5(2.3)
				& 80.3(2.3) & \textbf{84.0}(1.7) & \textbf{90.9}(1.7)
				& 29.5(7.0) & 86.6(4.5) & 85.8(4.7)
				& 65.9(6.5) & 92.3(0.6) & 93.4(2.2) & 76.2 \\
	CORAL 		& 80.3(1.9) & 63.8(3.1) & 62.1(3.0)
				& 77.6(1.2) & 61.2(2.4) & 64.3(2.9)
				& 66.6(2.2) & 69.1(2.6) & 82.8(2.8)
				& 72.0(1.7) & 74.2(2.2) & 89.6(1.6) & 72.0 \\
	TAISL 		& 90.0(1.9) & 85.3(3.1) & \textbf{90.6}(1.9)
				& 80.1(1.4) & 77.9(2.6) & 85.1(2.2)
				& 82.6(2.2) & 85.6(3.5) & 97.7(1.5)
				& \textbf{84.0}(1.0) & 87.6(2.1) & 95.9(1.0) & 86.9\\
	JGSA 		& 87.0(0.8) & 69.4(6.7) & 77.29(7.0)
				& 79.6(1.2) & 67.8(4.8) & 76.27(6.1)
				& 81.4(1.0) & 87.1(0.7) & 96.9(1.8)
				& 82.2(0.7) & 88.5(0.8) & 94.9(1.0) & 82.1 \\
	\hline
	SDTDL 		& \textbf{94.8}(3.2) & \textbf{89.5}(4.4) & 90.4(4.7)
				& \textbf{86.4}(2.5) & 82.8(5.7) & 88.8 (3.6)
				& \textbf{84.4}(2.2) & \textbf{91.7}(3.6) & \textbf{97.9}(1.7)
				& 83.9(1.1) & \textbf{92.1}(1.4) & \textbf{98.1}(1.2) & \textbf{90.1}\\
	\hline
	\end{tabular}
	}
% 	\end{center}
	% \label{tab:Accy_OC10_routine}
			\vspace{-0.3cm}
	\end{table*}

\noindent\textbf{Recognition Accuracy.}
Table \ref{tab:Accy_OC10_routine} shows that SDTDL achieves the highest accuracy in $8$ pairs out of $12$ and gains performance improvements in average accuracy of $3\%$ compared to the best method for comparison.
We observe that in C$\rightarrow$D and D$\rightarrow$C, our method reaches a close second to the best results ($90.4\%$ vs. $90.6\%$ and $83.9\%$ vs. $84.0\%$, respectively). The leading performance of SDTDL compared with other vector-based UDA methods indicates that the internal information of high-dimensional visual data are indeed crucial to cross-domain recognition.
It meanwhile demonstrates that SDTDL indeed effectively preserves the useful internal information in the visual data. We also observe that SDTDL outperforms TAISL in all the $12$ pairs, which demonstrates the proposed method is able to restrain the interference of domain factors and facilitate the discriminativeness of feature.
In Table \ref{tab:Accy_digit}, we can see that SDTDL outperforms both the competitive shallow and deep UDA methods on digit datesets. One one hand, this demonstrates the strong power for discriminative domain-invariant feature extraction of SDTDL. One the other hand, the results validate the advantageous over other methods of SDTDL when large training samples are unavailable for cross-domain recognition.

\noindent\textbf{Small Sample Size Scenarios.}
We evaluate the performance of SDTDL in addressing the small sample size problem through cross-domain recognition tasks W$\rightarrow$C and MNIST$\leftrightarrow$USPS.
For W$\rightarrow$C, $1{\sim}8$ random samples per class from domain W and all the target samples of domain C are selected to compose the dataset.
As shown in Fig.\ref{fig:visualization}~(e-f), SDTDL outperforms other three methods when the label source samples are limited, suggesting that SDTDL can achieve knowledge between domains when few label samples are available.
% It also should be pointed out that SDTDL underperforms when only one source sample from each class is available.
We also note that SDTDL underperforms when only one source sample from each class is available.
The reason is that the class mean of source sample becomes to zero in this case, thwarting the discriminative term in mode (\ref{eq:model}). In addition, we select all the source samples and $k{\in} \{1,2,4,8,16,32,64\}$ target samples per class to simulates the scenario of small sample size in target domain. Fig.~\ref{fig:visualization}~(f) shows that SDTDL offers advantages over other three competitive shallow methods when the number of target samples is limited.
For MNIST$\leftrightarrow$USPS, $k{\in}\{5,10,15,20\}$ random source samples per class and all the target samples are selected to compose the dataset.
Fig.~\ref{fig:visualization}~(g-h) show that SDTDL outperforms other three competitors when source samples are scarce in cross-domain digit recognition. Besides, the advantage of SDTDL over TAISL in recognition accuracy demonstates that the structured discrimination dictionary learning strategy of SDTDL can effectively address the small sample size problem in cross-domain recognition.

%------------------------Accuracy for digit recognition --------------------
\begin{table}[!tbp]  \small
	\caption{Accuracy($\%$) on Digit Dataset.}
	\label{tab:Accy_digit}
% 	\begin{center}
	\centering
	\resizebox{0.48\textwidth}{!}{
	\begin{tabular}{|l | c c c c c c c c c c|}
	\hline
	Method 	& TCA 	& GFK
			& SA 	 & JDA  & CORAL
			& TAISL 	& JGSA
			& DANN   & DDC	& SDTDL \\
	\hline
	$M \rightarrow U$ & 56.3	& 61.2
					& 67.8 & 67.2 &83.6
                    & 83.0 & 82.3 	
                    & 77.1 & 79.1  & \textbf{90.7}	\\
	$U \rightarrow M$ & 51.2	& 46.5
					& 48.8 & 59.7 & 78.5
                    & 82.6 & 87.8 	
                    & 73.0 & 66.5  & \textbf{89.1}	\\
    \hline
    MEAN  & 53.8	& 53.9
    & 58.3 & 63.4 & 81.1
    & 82.8 & 85.1
    & 75.1 & 72.8  & \textbf{89.9}	\\
	\hline
    \end{tabular}
    }
% 	\end{center}
	% \label{tab:Accy_OC10_routine}
			\vspace{-0.2cm}
	\end{table}

\noindent\textbf{Parameter Sensitivity Analysis.}
We investigate the parameter sensitivity of SDTDL w.r.t target domain weighting parameter $\theta$, parameter of intra-class variance $\lambda$ and target sample selection parameter $\delta$. Fig. \ref{fig:visualization} (i-j) validate that SDTDL achieves stable performance for a wide range of parameter settings for $\theta$ and $\lambda$.
The observation from  Fig.~\ref{fig:visualization}~(k) is two-folds: (1) a large proportion of target samples should be selected in SDTDL to insure the samples from each category are provided for training; (2) the proportion should be controlled within a certain range to prevent the negative effects of false labels.

\noindent\textbf{Convergence Analysis.}
We evaluate the convergence property of SDTDL by checking the prediction accuracy of target samples in each iteration.  Fig.~\ref{fig:visualization}~(l) shows the increasement of prediction accuracy along with dictionary learning process, indicating that the dictionary becomes more and more transferable and discriminative. This also demonstrate the effectiveness of our pseudo-label selection strategy in model training. Besides, we observe that dictionary evolution can reach the balance between domain-robust and discriminativeness within $10$ iterations in most cases.

\noindent\textbf{Dictionary Property Analysis.}
To demonstrate the efficacy of the learned domain-specific and class-specific sub-dictionaries in extracting domain information and class information, we analysis the reconstructed samples associated with the two sub-dictionaries. Concretely, we apply SDTDL to domain adaptation from MNIST to USPS (M$\rightarrow$U) and compare the original images and the domain-specific and class-specific reconstructed images. From the results in Fig.~\ref{fig:reconstruction}, we observe that the images in (b) and (e) contain more domain information, \emph{e.g.}, the light typeface style of MNIST and the boldface style of USPS, than the category information of digits.
We also note that the images in (d) and (f) contain far more category information than typeface information. The results demonstrate that the sub-dictionaries learned by SDTDL can focus on the domain factor and extract class information from data separately in domain-shift situation.
% %------------------Figure: Algorithm Property
% \begin{figure}[!tbp] \small
% \setlength{\abovecaptionskip}{0pt}
% \centering
% \includegraphics[width=1\linewidth]{alg_property.pdf}
% \caption{Investigate the property of SDTDL. (a) and (b) samll sample size.(c) convergence . (d)(e)(f) parameter sensitivity. (Better view zoomed in and with color)}
% \label{fig:alg_property}
% 		\vspace{-0.4cm}
% \end{figure}

% \vspace{-0.2cm}
\section{Conclusion}
% \vspace{-0.1cm}
Previous unsupervised domain adaptation methods vectorize multi-dimensional data in advance, leading to the loss of internal information which is critical to visual recognition applications. Besides, most existing methods are based on the assumption of plenty samples, which is rarely hold in practice. In this paper, we propose to learn a structured discriminative dictionary using tensor model. The dictionary is composed of multi-linear factor matrices, providing the capability to represent tensors. Moreover, domain-specific information and class-specific information of the cross-domain samples are depicted by the corresponding sub-dictionaries respectively. Our method shows strong power of feature extraction through knowledge transfer between domains, not only in traditional domain adaptation setting, but also in the setting of limited samples, which is rarely explored.

\clearpage
\small
\bibliographystyle{named}
\bibliography{ampc_ref}

\appendices
\section{Proof of Theorem 1}
The proof proof to Theorem \ref{th:th1} in the main paper is presented in this section. Theorem \ref{th:th1} provide the solution to the following optimization problem
%****************************************
\begin{equation}
	\begin{aligned}
		\underset{_{\mathbf{W}_{c}}}
		{\textmd{arg min}} & \; \left \{
		\begin{aligned}
			  & \|\mathcal{\widetilde{X}}^{c} - \llbracket \mathcal{A}_{c}^{c};\mathbf{W}_{c} \rrbracket \|_{F}^{2}
			+ \theta \|\mathcal{\widetilde{Y}}^{c} - \llbracket \mathcal{B}_{c}^{c};\mathbf{W}_{c} \rrbracket \|_{F}^{2} \\
			+ & \lambda \left ( \|\mathcal{A}_{c}^{c} - \mathcal{\bar{M}}_{c}^{B} \|_{F}^{2}
			+ \|\mathcal{B}_{c}^{c} - \mathcal{\bar{M}}_{c}^{A} \|_{F}^{2} \right ) \\
		\end{aligned}
		\right \} \\
		% s.\;t.             & \; \mathcal{A}_{c}^{c} \in R^{J_1\times J_2 \times \ldots \times J_M},
		% \; \mathcal{B}_{c}^{c} \in R^{J_1\times J_2 \times \ldots \times J_M}, \\
		\text{s.\;t.}  & \; W_c^{(m)T}W_c^{(m)}={\rm I}, \quad m=1,\ldots,M.	
	\end{aligned}
	\label{eq:submodel_AcWc}
\end{equation}

\begin{theorem}
\label{th:th1}
Let $\mathcal{\widetilde{Z}}^{c}=[\mathcal{\widetilde{X}}^{c}|\mathcal{\widetilde{Y}}^{c}]$ be the augmented sample tensor of class $c$ which is generated by concatenating $\mathcal{\widetilde{X}}^{c}$ and $\mathcal{\widetilde{Y}}^{c}$ along with the $M+1$-th mode. %Similarly, denote by $\mathcal{G}^{c}=[\mathcal{A}^{c}_c | \mathcal{B}^{c}_c]$ the augmentation of represenation.
Define matrix $\Phi$ as
%****************************************
\begin{equation}
	\Phi = \left(
	\begin{array}{cc}
			(1-\sqrt{\lambda}){\rm I}_{N_s^c}                    & \frac{\sqrt{\lambda}}{N_s^c}1_{N_s^c}1^{T}_{N_t^c} \\
			\frac{\sqrt{\lambda}}{N_t^c}1_{N_t^c}1^{T}_{N_s^c} & (\sqrt{\theta}-\sqrt{\lambda}){\rm I}_{N_t^c}
		\end{array}\right),
\label{eq:def_Phi}
\end{equation}
%****************************************
where ${\rm I}_{N_s^c}$ and ${\rm I}_{N_t^c}$ are identical matrices, $\mathbf{1}_{N_s^c}$ and $\mathbf{1}_{N_t^c}$ are the column vectors with all ones. Let $G_c^{(m)}$ be the mode-$m$ flatting matrix of $\mathcal{G}_{c} = \llbracket \mathcal{\widetilde{Z}}^{c};W^{(1)T}_c,\ldots, W^{(m-1)T}_c, {\rm I}, W^{(m+1)T}_c,\ldots,W^{(M)T}_c, \Phi \rrbracket$.
Then, the optimal $\mathbf{W}_c$ in (\ref{eq:submodel_AcWc}) is provided by $W_{c}^{(m)}=[\eta_{1}^{(m)},\ldots,\eta_{J_m}^{(m)}]$, with columns as the eigen-vectors corresponding to the first largest $J_m$ eigenvalues of the following eigenvalue-problem
%****************************************
\begin{equation}
% 	\widetilde{G}^{c,m}_{(m)}\big(\widetilde{G}^{c,m}_{(m)}\big)^T\eta = \alpha\eta
G_c^{(m)}\big(G_c^{(m)}\big)^T\eta = \alpha\eta.
	\label{eq:eigen-value}
\end{equation}
%****************************************
\end{theorem}
\;

\;

\;

\begin{proof}
Based on the formula (4.3) (4.4) in \cite{Kolda_SIAMReview09}, we have
%----------------eq:Obj1_1-------------------------------
\begin{equation}
\begin{array}{ll}
&    \begin{array}{rl}
    \underset{_{\mathbf{W}_{c}}}
		{\textmd{arg min}}  &
		\left\|\mathcal{\widetilde{X}}^{c} - \llbracket \mathcal{A}_{c}^{c};\mathcal{W}_{c} \rrbracket \right\|_{F}^{2}
			\\
			\text{s.\;t.}  & W_c^{(m)T}W_c^{(m)}={\rm I},\;m=1,\ldots,M
    \end{array} \\
\Longleftrightarrow & \\
 &
    \begin{array}{rl}
    \underset{_{\mathbf{W}_{c}}}
		{\textmd{arg min}}  &
		\left\|\llbracket \mathcal{A}_{c}^{c};\mathcal{W}_{c}^T \rrbracket \right\|_{F}^{2}
			\\
			\text{s.\;t.}  & W_c^{(m)T}W_c^{(m)}={\rm I},\;m=1,\ldots,M.
    \end{array}
\end{array}
\label{eq:obj1_1}
\end{equation}
%---------------------------------------------------------
Similarly, we have we have
%----------------eq:obj1_2-------------------------------
\begin{equation}
\begin{array}{ll}
&    \begin{array}{rl}
    \underset{_{\mathbf{W}_{c}}}
		{\textmd{arg min}}  &
		\left\|\mathcal{\widetilde{Y}}^{c} - \llbracket \mathcal{B}_{c}^{c};\mathbf{W}_{c} \rrbracket \right\|_{F}^{2}
			\\
			\text{s.\;t.}  & W_c^{(m)T}W_c^{(m)}={\rm I},\;m=1,\ldots,M
    \end{array} \\
\Longleftrightarrow & \\
 &
    \begin{array}{rl}
    \underset{_{\mathbf{W}_{c}}}
		{\textmd{arg min}}  &
		\left\|\llbracket \mathcal{\widetilde{Y}}^{c};\mathbf{W}_{c}^T \rrbracket \right\|_{F}^{2}
			\\
			\text{s.\;t.}  & W_c^{(m)T}W_c^{(m)}={\rm I},\;m=1,\ldots,M.
    \end{array}
\end{array}
\label{eq:obj1_2}
\end{equation}
%---------------------------------------------------------
Define $\mathcal{\widetilde{Z}}^{c}=[\mathcal{\widetilde{X}}_c^{c}|\mathcal{\widetilde{Y}}_c^c] \in\mathbb{R}^{J_1 \times J_2 \times \ldots \times J_M\times(N_s^c+N_t^c)}$, we can get the following equivalence with formula derivation
%----------------eq:obj2-------------------------------
\begin{small}
\begin{equation}
\begin{array}{ll}
\begin{array}{rl}
    \underset{_{\mathbf{W}_{c}}}
		{\textmd{arg min}}  &
		 \left\|\mathcal{A}_{c}^{c} - \mathcal{\bar{M}}_{c}^{B} \|_{F}^{2}
+ \|\mathcal{B}_{c}^{c} - \mathcal{\bar{M}}_{c}^{A} \right\|_{F}^{2}
			\\
			\text{s.\;t.}  & W_c^{(m)T}W_c^{(m)}={\rm I},\;m=1,\ldots,M
    \end{array}  & \\
\Longleftrightarrow & \\
   \begin{array}{rl}
    \underset{_{\mathbf{W}_{c}}}
		{\textmd{arg max}}  &
	\left\| \left \llbracket \mathcal{\widetilde{Z}}^{c} \times_{M+1}
    \left(
    \begin{bmatrix}
\begin{smallmatrix}
		{\rm I}_{N_s^c}
		& \frac{-1}{N_s^c}\mathbf{1}_{N_s^c}\mathbf{1}^{T}_{N_t^c} \\
			\frac{-1}{N_s^c}\mathbf{1}_{N_t^c}\mathbf{1}^{T}_{N_s^c}
		& {\rm I}_{N_t^c}
		\end{smallmatrix}
\end{bmatrix}
% 	\begin{array}{cc}
% 		{\rm I}_{N_s^c}
% 		& \frac{-1}{N_s^c}\mathbf{1}_{N_s^c}\mathbf{1}^{T}_{N_t^c} \\
% 			\frac{-1}{N_s^c}\mathbf{1}_{N_t^c}\mathbf{1}^{T}_{N_s^c}
% 		& {\rm I}_{N_t^c}
% 		\end{array}
	\right) ; \mathbf{W}_c \right \rrbracket  \right\|_{F}^2
			\\
			\text{s.\;t.}  & W_c^{(m)T}W_c^{(m)}={\rm I},\;m=1,\ldots,M.
    \end{array} &
\end{array}
\label{eq:obj2}
\end{equation}
	\end{small}
%---------------------------------------------------------
Taking (\ref{eq:obj1_1})(\ref{eq:obj1_2})(\ref{eq:obj2}) into account, the optimal problem (\ref{eq:model}) is equivalent to the following optimal problem
%****************************************
\begin{equation}
	\begin{array}{rl}
		\underset{_{\mathbf{W}_{c}}} {\textmd{arg max}} &
		\left\|  \mathcal{\widetilde{Z}}^{c}\Phi \right|_F^2 \\
		\text{s.\;t.}  &  W_c^{(m)T}W_c^{(m)}=\rm I, \;m=1,\ldots,M.	
	\end{array}
	\label{eq:newmodel_1}
\end{equation}
%****************************************

For enable better readability, we define intermediate variable
\begin{equation}
\mathcal{G}_{c} = \llbracket \mathcal{\widetilde{Z}}^{c};W^{(1)T}_c,\ldots, W^{(m-1)T}_c, {\rm I}, W^{(m+1)T}_c,\ldots,W^{(M)T}_c, \Phi \rrbracket
\end{equation}
So far, we can obtain the optimal factor matrix $W_c^{(m)}$ for each mode $m$ by solving the following optimal problem
%****************************************
\begin{equation}
	\begin{array}{rl}
		\underset{_{W_c^{(m)}}} {\textmd{arg max}} &
		\left\| W_c^{(m)}G_c^{(m)} \right\|_F^2 \\
		\text{s.\;t.}  &  W_c^{(m)T}W_c^{(m)}={\rm I},
	\end{array}
	\label{eq:newmodel_2}
\end{equation}
%****************************************
% where $G_c^{(m)}$ is the mode-$m$ flatting matrix of $\mathcal{G}_c$. According to Lagrange multiplier method, the optimal solution of (\ref{eq:newmodel_2}) is $W_{c}^{(m)}=\left[\eta_{1}^{(m)},\ldots,\eta_{J_m}^{(m)}\right]$ with colums as the eigen-vectors corresponding to the first largest $J_m$ eigenvalues of the following eigenvalue-problem
% %****************************************
% \begin{equation}
% 	\label{eq:eigen-value}
% 	G_c^{(m)}\big(G_c^{(m)}\big)^T\eta = \alpha\eta
% \end{equation}
% %****************************************
where $G_c^{(m)}$ is the mode-$m$ flatting matrix of $\mathcal{G}_c$. According to Lagrange multiplier method, the optimal solution of (\ref{eq:newmodel_2}) is $W_{c}^{(m)}=\left[\eta_{1}^{(m)},\ldots,\eta_{J_m}^{(m)}\right]$ with columns as the eigen-vectors corresponding to the first largest $J_m$ eigenvalues of eigenvalue-problem (\ref{eq:eigen-value}).
\end{proof}

\section{Dictionary Property Analysis}
\begin{figure}[pt]\small
	\centering
	\includegraphics[width=1\linewidth]{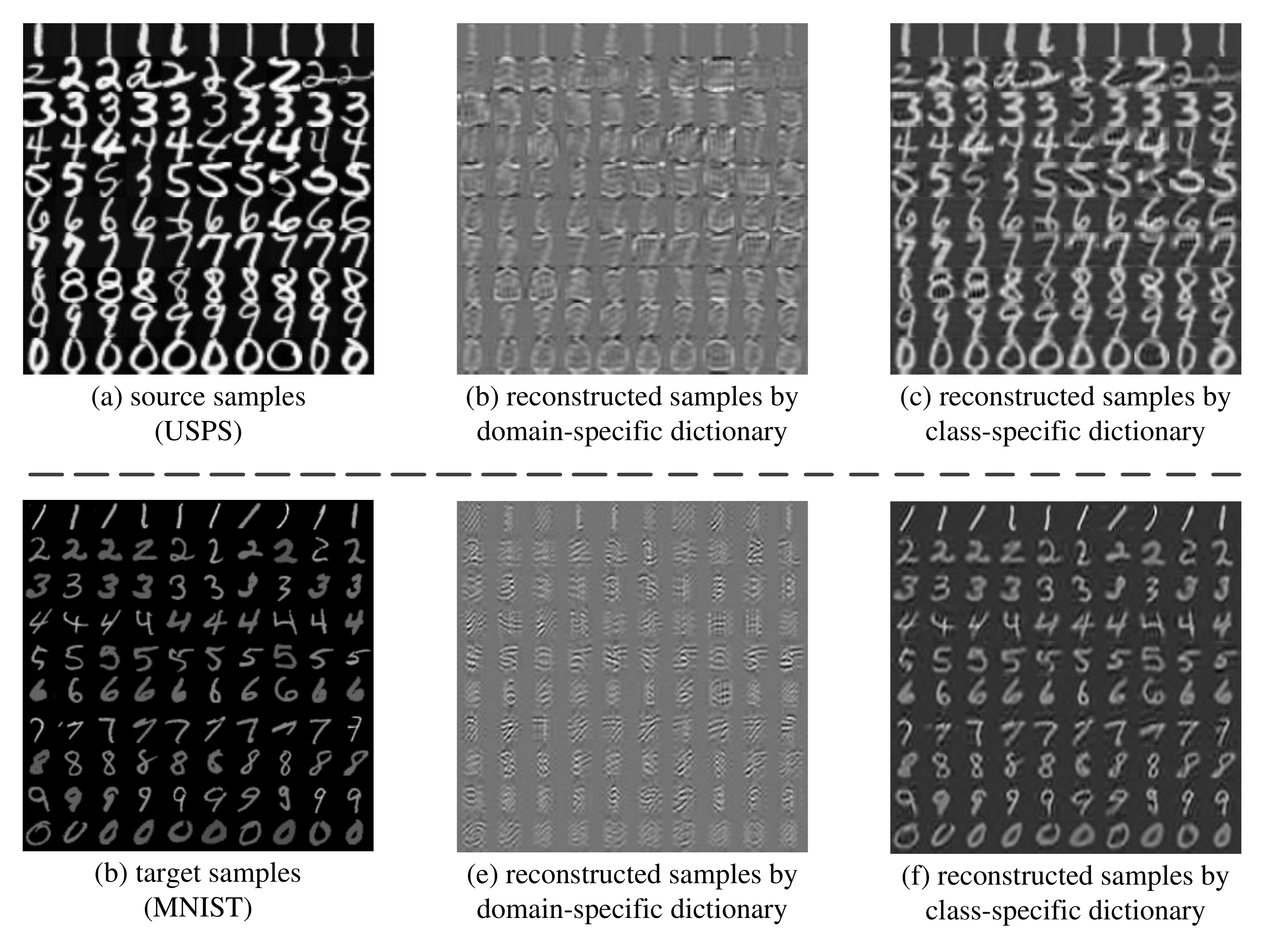}
	\caption{Visualization of reconstructed samples by domain-specific dictionary and class-specific dictionary respectively in the task USPS $\rightarrow$ MNIST.}
	\label{fig:rec_u2m}
	%\vspace{-0.4cm}
\end{figure}

In this section, we provide additional experimental results to demonstrate the efficacy of the learned domain-specific and class-specific sub-dictionaries in extracting domain information and class information.We apply SDTDL to the task of transferring from USPS to MNIST (U$\rightarrow$M), in which we compare the original images and the domain-specific and class-specific reconstructed images. The results in Fig. \ref{fig:rec_u2m} demonstrate that the domain-specific sub-dictionary and the class-specific sub-dictionary learned by SDTDL are able to extract the domain information and class information from cross-domain data respectively.

\end{document}